\documentclass{article}

\usepackage{PRIMEarxiv}

\usepackage[utf8]{inputenc} 
\usepackage[T1]{fontenc}    
\usepackage{hyperref}       
\usepackage{url}            
\usepackage{booktabs}       
\usepackage{amsfonts}       
\usepackage{nicefrac}       
\usepackage{microtype}      
\usepackage{lipsum}
\usepackage{fancyhdr}       
\usepackage{graphicx}       
\graphicspath{{media/}}     
\usepackage{listings}
\usepackage{xcolor}

\title{SynChart: Synthesizing Charts from Language Models
}

\author{
  Mengchen Liu, Qixiu Li, Dongdong Chen, Dong Chen, Jianmin Bao, Yunsheng Li \\
  Microsoft \\
  \texttt{\{mengcliu, v-qixiuli, dongdong.chen, doch, jianmin.bao, yunshengli\}email@microsoft.com} \\
}

\begin{document}
\maketitle

\begin{abstract}
With the release of GPT-4V(O), its use in generating pseudo labels for multi-modality tasks has gained significant popularity. 
However, it is still a secret how to build such advanced models from its base large language models (LLMs).
This work explores the potential of using LLMs alone for data generation and develop competitive multi-modality models focusing on chart understanding.
We construct a large-scale chart dataset, SynChart, which contains approximately 4 million diverse chart images with over 75 million dense annotations, including data tables, code, descriptions, and question-answer sets. 
We trained a 4.2B chart-expert model using this dataset and achieve near-GPT-4O performance on the ChartQA task, surpassing GPT-4V.
\end{abstract}


\section{Introduction}

Since the release of GPT-4V(O), using them to generate pseudo labels for multi-modality tasks has become more and more popular~\cite{chen2023sharegpt4v}
While we often "stand on the shoulders of giants," the process of building the giant itself—specifically, constructing GPT-4V(O) from its foundational large language model (LLM), GPT-4—remains a mystery.
In this work, we explore the potential of using LLMs alone to build a competitive multi-modality model.
Given budget constraints, we focus on a specific domain—chart understanding—rather than building a general multi-modality model.

Since the quantity and quality of data are key determinants of model performance, this work focuses on building a large-scale chart dataset and applying well-established training pipelines. There are two major challenges in constructing such a dataset: first, collecting a diverse set of chart images, and second, the more critical and difficult task of obtaining high-quality labels for these images. 
To address these challenges, we explored various dataset-building alternatives and carefully analyzed the trade-offs between data quality and quantity.
This analysis led us to our data generation approach: synthesizing data from LLMs.

\begin{figure}[h]
  \centering
  {\includegraphics[width=0.8\linewidth]{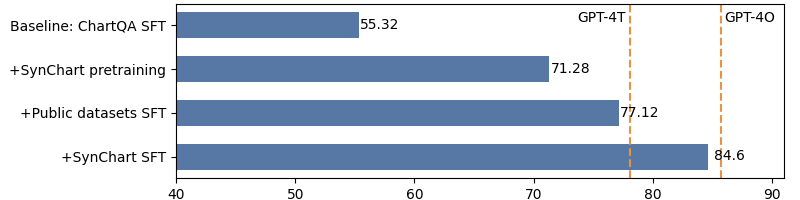}}
  \caption{ChartQA accuracy highlighting different component contributions.}
  \label{fig:teaser}
\end{figure}

Using this scalable data generation process, we developed a large-scale chart dataset called SynChart, which contains approximately 4 million diverse chart images. Each image is accompanied by rich annotations, including the underlying data table, the code used to generate the chart, detailed descriptions, and a set of questions and answers. 
The dataset card is shown in Table.~\ref{tab:card}.
Leveraging SynChart, we trained a chart-specific multi-modality model by combining Phi3.5 (3.8B) and CLIP-L (0.3B). 
As illustrated in Figure~\ref{fig:teaser}, the model's performance on the ChartQA benchmark is close to GPT-4O and surpasses GPT-4V.

\section{Design Rationales}
\label{sec:design}

Our primary objective is to create a large-scale chart dataset specifically for training multi-modality models. This dataset must include two key components: (1) a substantial collection of chart images, and (2) high-quality annotations that accompany these images.

\begin{table}[h]
 \caption{Comparison of data collection approaches. In column ``No. Charts'', K denote thousand, M million and B denotes billion.}
  \centering
  \begin{tabular}{llll}
    \toprule
    Approaches & Example Datasets & No. Charts  & Label Quality \\
    \midrule
    Collecting from domains with chart images & Chart2Text & K-M & High \\
    Filtering from general image datasets & N.A. & M-B & Medium \\
    Synthesizing from code & ChartLLAMA & M-B & High \\
    \bottomrule
  \end{tabular}
  \label{tab:data-collection-alter}
\end{table}

\noindent{\textbf{Design Alternatives.}} To collect chart images and corresponding labels, previous literature proposed two main approaches. 

One approach is to collect chart images from specific domains.
For example, Chart-to-text~\cite{kantharaj2022chart} gathered approximately 30K chart images from two sources:: Pew~\cite{pew} and Statista~\cite{statista}.
The advantage of this method is the high-quality associated labels, such as human-written summaries and human-annotated data tables. However, the limitation lies in the relatively small number of available chart images. Based on our estimates, fewer than 1 million charts are accessible across the websites we surveyed. For example, Gallup~\cite{gallup} offers around 80K chart images.

The second approach involves synthesizing chart images using chart-drawing libraries like Matplotlib. For example, ChartLLAMA~\cite{han2023chartllama} generated 11K chart images by leveraging GPT-4’s coding capabilities to produce Matplotlib code. This method has the potential to scale well beyond millions of chart images. Moreover, obtaining high-quality labels for these generated images is straightforward, as the ground truth data is inherently available in the code. The main challenge with this approach, however, is ensuring that the generated charts are both representative and diverse.

In addition to the two approaches mentioned earlier, we can also extract chart images from large-scale general image-text interleaved datasets, such as Obelics~\cite{laurenccon2024obelics} and MINT-1T~\cite{awadalla2024mint}. We started with Obelics to assess the potential data scale and label quality. To extract chart images, we developed a cascaded image classifier composed of three consecutive sub-classifiers: (1) a classifier to filter out natural images, (2) a classifier to remove non-chart images, and (3) a final classifier to exclude "science charts" like waveform diagrams. We opted for a cascaded approach rather than a single classifier due to its higher precision (96.7\% on our held-out, human-annotated test set, compared to less than 90\% with a single classifier). As a result, we obtained approximately 3 million chart images from Obelics.

In image-text interleaved datasets, surrounding text naturally serves as labels for chart images. To assess the quality of these labels, we randomly selected 100 chart images and manually annotated the relevant text from the two surrounding paragraphs. An example is shown in Figure~\ref{fig:chart-sample-obelics}
. Our analysis revealed a weak correlation (less than 50\% token-level relevance) between the chart images and their surrounding paragraphs in the samples we examined.

Table~\ref{tab:data-collection-alter} summarizes the trade-offs between quality and quantity for the aforementioned approaches.
Based on this analysis, we have chosen synthesized chart data as our primary data source due to its scalability. Additionally, we have integrated publicly available datasets to leverage their high-quality labels.

\section{SynChart Dataset}
\label{sec:data}

\begin{figure}
  \centering
  {\includegraphics[width=0.9\linewidth]{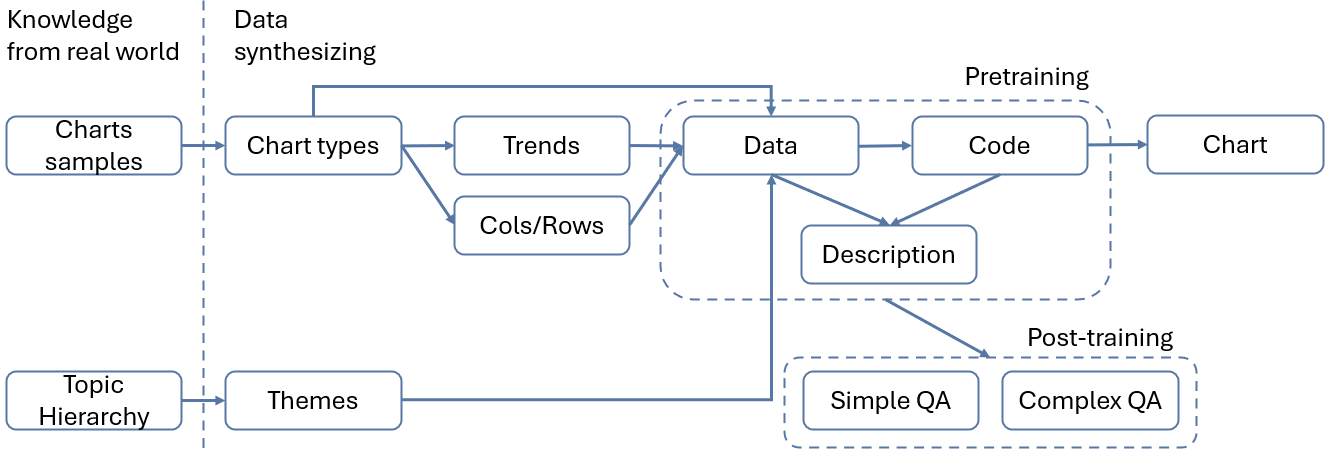}}
  \caption{Data generation pipeline.}
  \label{fig:pipeline}
\end{figure}


\subsection{Dataset Building}

In multi-modality model training, the primary requirements for the training dataset are scale, diversity, representativeness, and label density. As outlined in Section ~\ref{sec:design}, we focus on synthesizing data to meet these needs. Our approach to generating pretraining data involves three key stages as shown in Figure~\ref{fig:pipeline}.

In Stage 1, we create a set of diverse data tables that accurately represent real-world scenarios. In Stage 2, we employ LLMs to generate suitable code for visualizing the data as chart images. Finally, in Stage 3, we utilize LLMs to produce a set of question-answer pairs for post-training.

\noindent{\textbf{Stage 1: Data Generation.}}
In this stage, we aim to generate a set of diverse data tables that accurately reflect real-world data. To achieve this, we first establish essential constraints tailored to different types of charts, as each chart type has its own specific requirements and is suited for various kinds of data.

We begin by identifying the common \textbf{types of charts} used in SynChart. To ensure alignment with real-world applications, we randomly select samples of chart images filtered from Obelics and ChartBench~\cite{xu2023chartbench}. We then label the chart types of these images through a combination of model predictions and human annotations. Based on the labeling results, we identify and focus on the top nine most common chart types (detailed in Appendix \ref{appx:types-of-charts}).

With the identified chart types, we can generate a diverse set of trends and column/row constraints tailored to each chart type. Utilizing large language models (LLMs) and guided by prompts detailed in Appendix~\ref{appx:trends-of-charts}, we generated a total of 74 trends for the nine chart types. For the column and row constraints, we manually refined the outputs from the LLM, and these constraints are presented in Table~\ref{tab:col-row-constraints}.

To ensure alignment with real-world data, we further generate themes for each data table based on topics from the widely used question-answering platform StackExchange~\cite{stackexchange}.
We chose this source because it provides detailed definitions for each topic, aiding the LLM in generating relevant data. Specifically, we crawled 105 topics from the platform and used the LLM to create approximately 10 themes for each topic, resulting in a total of 1,262 themes.

By combining all this information, we employ the LLM to generate a diverse set of data tables as well as a detailed description of the data table. 
The specific prompts used in this process can be found in Appendix~\ref{sec:appx-prompt-data}.

\noindent{\textbf{Stage 2: chart generation.}}
Based on the generated data tables, we employ a variety of data visualization engines to produce the code for drawing charts. We select these engines according to two main criteria: (1) the LLM's demonstrated ability to generate code using each engine, and (2) the engine's capacity to create a wide range of chart types relevant to SynChart. As a result, we choose four commonly used engines: Matplotlib, Seaborn, Plotly, and Bokeh. The chart types supported by each engine are listed in Table~\ref{tab:engine-chart-types}.

Combining the available engines, chart types, and generated data tables, we use the LLM to produce code snippets for creating the corresponding charts. These code snippets are then executed to generate the chart images. Throughout this process, we employ a human-in-the-loop approach to correct common errors in the code snippets. This iterative refinement leads to an increase in the success rate of generating chart images, rising from 64.0\% to 76.8\% for Matplotlib.
This iterative refinement process enabled us to collect approximately 600K more chart images.

\noindent\textbf{Stage 3: question answer pair generation.}
To prepare SynChart for post-training use, we utilize LLMs to generate a set of question-and-answer pairs for each chart image. Specifically, we create two types of questions. The first type consists of simple questions that require only a single word or phrase as the answer. The second type includes more complex questions that necessitate a reasoning process to arrive at the correct answer.

\subsection{Dataset Card}
\label{sec:card}

The dataset card is shown in Table~\ref{tab:card}. An example from the dataset can be found in Appendix~\ref{appx:sample}.

\begin{table}[h]
 \caption{Dataset Card}
  \centering
  \begin{tabular}{lllll}
    \toprule
    Annotation Type & No. Images & No. Annotations & Format \\
    \midrule
    Data table      & 3.93M     & 3.93M          & csv    \\
    Code            & 3.93M     & 3.93M          & python \\
    Descriptions    & 3.93M     & 7.86M          & text    \\
    Simple question answers & 2.51M & 45.8M       & QA pair \\
    Complex question answers & 2.02M & 13.9M       & QA pair \\
    \bottomrule
  \end{tabular}
  \label{tab:card}
\end{table}

\section{Experiments}
\label{sec:exp}

\noindent{\textbf{Settings}.} Since our focus in this work is on data building, the training pipeline follows a well-established framework~\cite{abdin2024phi}.
Specifically, we utilize Phi-3.5-mini-instruct (3.8B) as the base LLM and CLIP ViT-L/14 as the vision encoder. The maximum image resolution is set at 1344×1344 pixels.

The training process is divided into two stages: pretraining and post-training. During pretraining, we utilize annotations that include code, data tables, and descriptions. In the post-training phase, we primarily rely on annotations of questions and answers. Additionally, we incorporate a collection of public datasets as another source of high-quality labels.

\subsection{Main Results}

As shown in Table~\ref{tab:main-results}, the model trained on SynChart achieves performance levels close to GPT-4O on the ChartQA~\cite{masry2022chartqa} benchmark, outperforming all public small models. Notably, the majority of improvements over ChartLlama come from human-created splits. Furthermore, when compared to large public models with over 70 billion parameters, our model remains competitive. These results underscore the effectiveness of the SynChart dataset.

\begin{table}[h]
 \caption{Main results.}
  \centering
  \begin{tabular}{lllll}
    \toprule
    Model & Model size & ChartQA-Human  & ChartQA-Augmented & ChartQA-Avg \\
    \midrule
    \textit{Proprietary} & & & & \\
    GPT-4O       & N.A.    &   -    &   -    & 85.7  \\
    GPT-4V       & N.A.    &   -    &   -   & 78.5  \\
    \midrule
    \textit{Public Large Models} & & & & \\
    LLaVA OneVision-72B     & 72B   &  - & - & 83.7  \\
    Llama 3-V 70B           & >70B   & - & - & 83.2 \\ 
    \midrule
    \textit{Public Small Models} & & & & \\
    Llama 3-V 8B & >8B    & - & - & 78.7 \\
    ChartLlama   & 7B & 48.96 & 90.36 & 69.7 \\
    LLaVA OneVision-7B     & 7B    &  - & - & 80.0  \\
    SynChart &  4.2B     &  74.24  &  94.96  & 84.6 \\
    \bottomrule
  \end{tabular}
  \label{tab:main-results}
\end{table}

\subsection{Ablations}

Table~\ref{tab:ablations} illustrates the contributions of each data component used in our training process. 
We use the ``Baseline*'' from ChartLlama as a strong reference point for our work.
The baseline model, with 7 billion parameters, is larger than our model and is pretrained on image captioning data, followed by fine-tuning on the ChartQA training split. 
From this starting point, we systematically add data components and evaluate their contributions. As shown in in Table~\ref{tab:ablations}, we achieve approximately a 30\% improvement in ChartQA performance. 
This enhancement is primarily attributed to pretraining and post-training with SynChart, while incorporating public datasets from similar domains also proves beneficial. 
Consequently, the construction of high-quality yet small-scale datasets remains essential for improving model performance and aligning with human understanding.

\begin{table}[h]
 \caption{Data components contributions. ``Human'', ``Augmented'', and ``Avg'' all refer to ChartQA accuracy.}
  \centering
  \begin{tabular}{llllll}
    \toprule
    Model & Pretraining & Post-training  & Human & Augmented & Avg \\
    \midrule
    Baseline & LLAVA & ChartQA & 37.68 & 72.96 & 55.32  \\
    +SynChart Pretraining & SynChart & ChartQA & 53.12 & 89.44 & 71.28 \\
    +Public post-training & SynChart & ChartQA + Public & 61.52 & 92.72 & 77.12 \\
    +SynChart Post-training & SynChart & ChartQA + Public + SynChart & 74.24 & 94.96 & 84.60 \\
    \bottomrule
  \end{tabular}
  \label{tab:ablations}
\end{table}

\subsection{Scaling Property}

To demonstrate the scaling properties of SynChart, we conduct two ablations focused on data and training costs. As shown in Table~\ref{tab:scaling}, utilizing more data from SynChart during post-training leads to improved performance. 
Notably, at the maximum training cost we tested, the model's performance has not yet plateaued. 
This finding addresses a critical concern regarding synthetic datasets: that models may quickly reach saturation due to limited data diversity.

\begin{table}[h]
 \caption{Scaling property of SynChart dataset. ``Ratio'' is the data ratio between public datasets (including ChartQA training split) and SynChart post-training data. The training cost of purely using public datasets is treated as 1.}
  \centering
  \begin{tabular}{lllll}
    \toprule
    Ratio & Training Cost & ChartQA-Human & ChartQA-Augmented & ChartQA-Avg   \\
    \midrule
    1:0   & 1             & 61.52 & 92.72     & 77.12  \\
    1:1   & 2             & 69.20 & 93.12     & 81.16  \\
    1:2   & 3             & 71.04 & 93.20     & 82.12  \\
    1:6   & 7             & 72.96  & 94.08 & 83.52 \\
    1:6   & 14            & 74.24 & 94.96 &  84.60   \\
    \bottomrule
  \end{tabular}
  \label{tab:scaling}
\end{table}

\section{Related Work}
\label{sec:related}

Among the various chart dataset building papers~\cite{han2023chartllama,masry2023unichart,lee2023pix2struct,liu2022matcha,liu2022deplot}, the most closely related work is ChartLlama~\cite{han2023chartllama}.
Both studies share the fundamental approach of using LLMs to generate synthetic chart data. However, this work offers two significant contributions.

First, we provide a detailed analysis of alternative data collection methods. The results of this analysis underscore the necessity of a synthesizing process in chart dataset construction. 
Second, we achieve a substantially larger data scale; for instance, SynChart contains over 300 times more chart images (3.93 million compared to 11,000 in ChartLlama). 
In the scaling process, we focused on maximizing data diversity by incorporating a range of chart metadata, including types, themes, data constraints, trends, and engines.

\section{Conclusion}

In this work, we construct a large-scale chart dataset, SynChart, by synthesizing data from LLMs. The dataset comprises approximately 4 million diverse chart images, accompanied by dense annotations, including data tables, code, descriptions, and question-answer sets. We trained a 4.2 billion parameter chart-expert model using this dataset, achieving performance levels close to GPT-4O on the ChartQA task.

Looking ahead, there are several research directions to enhance the quality of the dataset. These include expanding the variety of chart types, filtering chart images based on visual quality, and incorporating support for dashboards with multiple sub-charts.

\bibliographystyle{unsrt}  
\bibliography{references}

\appendix

\section{Data Example}
\label{appx:sample}

Here we show one example used in the training.

\begin{figure}[b]
  \centering
  {\includegraphics[width=0.7\linewidth]{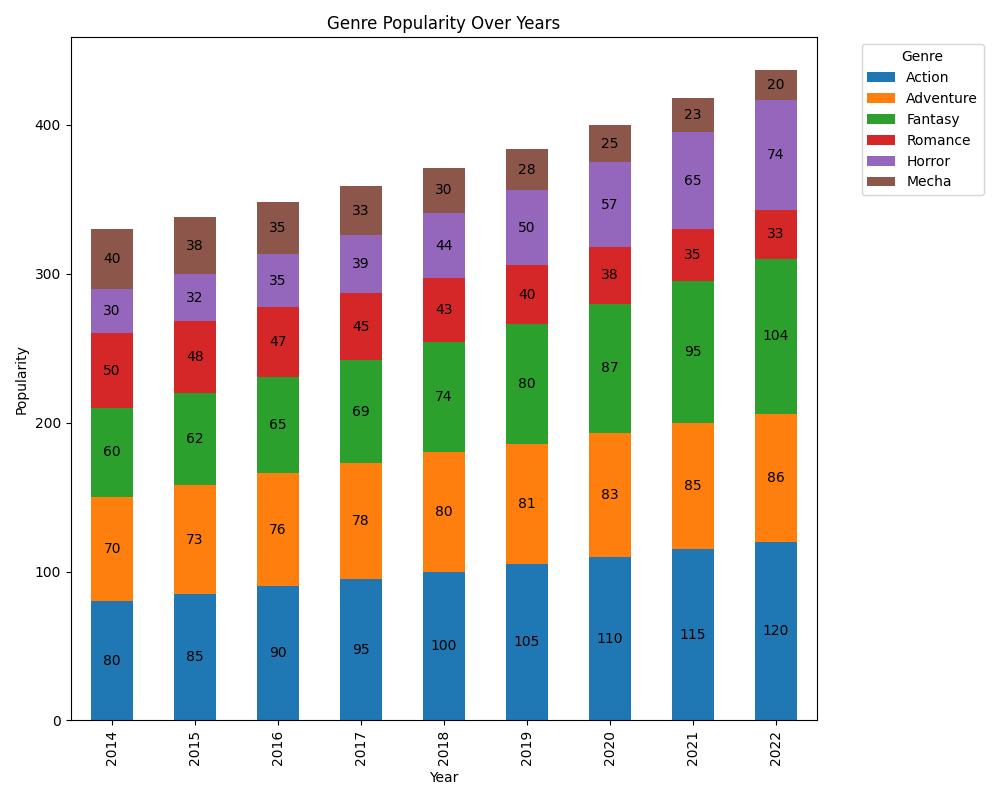}}
  \caption{Chart image sample used in pretraining.}
  \label{fig:chart-sample-pretraining}
\end{figure}

\textbf{Chart image}:

The chart image is shown in Figure~\ref{fig:chart-sample-pretraining}.

\textbf{Data table}:

Genre,2014,2015,2016,2017,2018,2019,2020,2021,2022

Action,80,85,90,95,100,105,110,115,120

Adventure,70,73,76,78,80,81,83,85,86

Fantasy,60,62,65,69,74,80,87,95,104

Romance,50,48,47,45,43,40,38,35,33

Horror,30,32,35,39,44,50,57,65,74

Mecha,40,38,35,33,30,28,25,23,20

\textbf{Code}:

\begin{lstlisting}[language=Python, caption=Example Python Code]

import numpy as np
import io
import matplotlib.pyplot as plt
import pandas as pd
import numpy as np

# Given CSV data
csv_data = """
Genre,2014,2015,2016,2017,2018,2019,2020,2021,2022
Action,80,85,90,95,100,105,110,115,120
Adventure,70,73,76,78,80,81,83,85,86
Fantasy,60,62,65,69,74,80,87,95,104
Romance,50,48,47,45,43,40,38,35,33
Horror,30,32,35,39,44,50,57,65,74
Mecha,40,38,35,33,30,28,25,23,20
"""

# Convert CSV data to DataFrame
data = pd.read_csv(io.StringIO(csv_data), index_col='Genre')

# Plotting
fig, ax = plt.subplots(figsize=(10, 8))
data.T.plot(kind='bar', stacked=True, ax=ax)

# Adding titles and labels
plt.title('Genre Popularity Over Years')
plt.xlabel('Year')
plt.ylabel('Popularity')

# Annotating data values
for container in ax.containers:
    ax.bar_label(container, label_type='center')

# Setting legend
plt.legend(title='Genre', bbox_to_anchor=(1.05, 1), loc='upper left')

# Save the figure
plt.tight_layout()
plt.savefig('1.jpg', format='jpg')

\end{lstlisting}

\textbf{Data description}:

The chart titled "Anime Genres and Their Unique Features" is a stacked bar plot designed to showcase the trends in popularity and production of different anime genres over the years from 2014 to 2022. The chart includes six distinct anime genres: Action, Adventure, Fantasy, Romance, Horror, and Mecha, each represented over a span of nine years.

The general trend observed from the chart data indicates a clear growth in the Fantasy and Horror genres, showcasing them as emerging categories within the anime industry. Specifically, the Fantasy genre has shown a remarkable increase from 60 in 2014 to 104 in 2022, and the Horror genre has grown from 30 in 2014 to 74 in 2022. This suggests a growing interest and production in these genres.

On the other hand, the Romance and Mecha genres are depicted as declining categories. Romance has steadily decreased from 50 in 2014 to 33 in 2022, and Mecha has seen a more significant drop from 40 in 2014 to 20 in 2022, indicating a possible shift in audience preferences away from these themes.

The Action and Adventure genres have shown moderate growth, with Action consistently growing from 80 in 2014 to 120 in 2022, and Adventure showing a slower pace of growth from 70 in 2014 to 86 in 2022. This steady growth suggests a stable interest in these genres, with Action being the most popular genre throughout the years.

Outliers in the data are particularly notable in the Fantasy and Horror genres, which have seen unexpected spikes in growth compared to other genres. These outliers suggest that certain years or events have significantly boosted the popularity and production of anime in these genres, possibly due to breakthrough titles or increased global interest.

In comparison, the decline in the Romance and Mecha genres is significant when juxtaposed with the growth of Fantasy and Horror, highlighting a shift in consumer interest and market dynamics within the anime industry. The data suggests that while traditional genres like Action and Adventure maintain a steady fanbase, emerging genres like Fantasy and Horror are capturing the imagination and interest of a broader audience, possibly due to evolving storytelling techniques and thematic exploration.

Overall, the chart provides a comprehensive overview of the shifting landscape of anime genres, highlighting emerging trends, declining interest in certain themes, and the presence of outliers that defy general trends. This analysis can be invaluable for content creators, marketers, and industry analysts looking to understand the evolving preferences of the anime audience.

\textbf{Chart description}:

The chart is a stacked bar plot illustrating the popularity of various genres over the years from 2014 to 2022. Each bar represents a year, and the segments within each bar denote the popularity of genres such as Action, Adventure, Fantasy, Romance, Horror, and Mecha. The y-axis measures the popularity, presumably in a quantifiable metric like sales, views, or ratings, while the x-axis lists the years.

The colors in the bars are distinct for each genre, making it easy to differentiate between them. The legend on the right side of the chart identifies which color corresponds to which genre. This layout allows viewers to track the popularity trends of each genre over time at a glance.

One can observe that the Action genre shows a consistent increase in popularity over the years, starting from 80 in 2014 and reaching 120 by 2022. Similarly, the Fantasy genre also shows a significant upward trend. In contrast, the Romance genre exhibits a gradual decline in popularity, starting from 50 in 2014 and decreasing to 33 by 2022. The Mecha genre, while starting at a lower point than others, also shows a decrease, highlighting a potential shift in audience preferences away from this genre.

This chart is valuable for stakeholders in the entertainment industry, such as producers, marketers, and content creators, as it provides insights into changing trends in genre popularity, which can inform future projects and marketing strategies.

\textbf{Simple question and answer}:

Question 1: What is the most popular genre in 2022?

Answer 1: Action

$...$

Question n: Which genre has decreased the most by 2022 from its 2014 value?

Answer n: Mecha

\textbf{Complex question and answer}:

Question: How much has the Horror genre grown from 2018 to 2022?

Answer: On the chart, the segment representing the Horror genre in 2018 shows a value of 44, which grows to 74 by 2022. The increase in the Horror genre's popularity from 2018 to 2022 can be calculated by subtracting the value in 2018 from the value in 2022. That calculation is 74 (2022) - 44 (2018) = 30.


\section{Chart Sample from Obelics}

Chart image sample from Obelics and its surrounding text. The relevant information is highted in bold

``[omitted 346 words]

\textbf{And before you chalk this up to being the rant of one disgruntled writer, check out Netflix’ ISP rankings for January.}

<image>

That’s right. \textbf{Out of the 17 companies ranked, Comcast finished 14th.} Looking for a way to get the most out of your Netflix subscription? Wanna sample their new 3D offerings? First try Google Fiber, Cablevision – Optimum, Cox, Suddenlink, Charter, Time Warner Cable (about to be Comcast!), Verizon – FIOS, Bright House, Windstream, Frontier, AT\&T – U-Verse, and Mediacom. Then, if that doesn’t work out for ya, OK, sure, you can give Comcast a shot. And wait, you didn’t just utter “4K” did you? Stop it.

So, um, yeah… what’re your thoughts on the proposed merger?''

\begin{figure}[h]
  \centering
  {\includegraphics[width=0.4\linewidth]{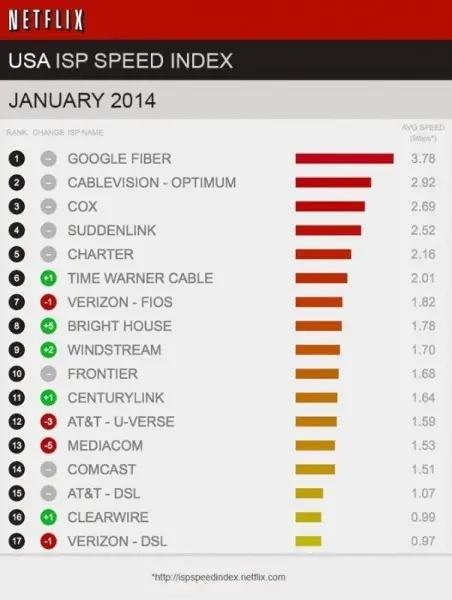}}
  \caption{Chart image sample from Obelics.}
  \label{fig:chart-sample-obelics}
\end{figure}

\section{Prompts}

\subsection{Prompt for Theme Generation}
\label{sec:appx-prompt-theme}

Generate a set of themes for this topic: Generate a set of themes for this topic: "\{topic\}", 
topic definiton: \{definition\} Each theme is a phrase. Return as json format.

\subsection{Prompt for Trend Generation}
\label{appx:trends-of-charts}

List possible trends found in this type of chart: \{chart\_type\}. Return as json format.

\subsection{Prompt for Col/Row Constraints Generation}

For a tabular data, to visualize it with a \{chart\_type\}, what is the suitable range of number of rows and columns?
Respond with conclusion first, followed by an explanation.

\subsection{Prompt for Data Generation}
\label{sec:appx-prompt-data}

You are an expert at generating data in csv format. You receive several key characteristics about the data. 
Your final output should include data in CSV format for the chart, and a comprehensive description of the chart data and figure.

\#\#Expected characteristics of the data in the chart.
The chart is \{chart\_type\}. 
The theme of the chart is \{theme\}. 
Different series of data in the chart can have different trends. The trends in the chart data should include as many of the given trends as possible: \{trends\}.
The data should be diverse and contain several outliers. The numbers of columns and rows should satisfy: \{row\_column\}. \{data\_constraints\}
You can list the nouns you know, which are related with the theme, along the first column and row of the table.

\#\# Requirement about the description
The description should focus on several key elements: the chart's theme, the general trend of the data, individual trends 
within the data, the comparison between data, and any outliers present in the chart. 

\#\# Requirement about the output
Your output should comprise the generated data wrapped in <data start> and <data end>, and detailed descriptions 
about the chart wrapped in <description start> and <description end>.

\subsection{Prompt for Code Generation}

You are a specialist in two aspects, drawing charts with \{engine\}, and providing detailed descriptions about the chart. 
You receive the data in the format of csv table. 
You need to generate Python code to plot the given data as a chart figure and providing detailed description about the figure.

Additional requirements:

You can freely set the chart styles to increase the diversity, including title, legend, labels on x-axis and y-axis. 
You can annotate data values above the point on the chart figure.

Do not use show function to show the figure.
Save the figure as a jpg file, with filename "\{filename\}" 
The csv data should be listed in the code. 

The output contains two parts.
The first part is the generated Python code wrapped in <code start> and <code end>. 
Next is the detailed description about the chart wrapped in <description start> and <description end>.
The code should be able to be executed without external files.

Draw a \{chart\_type\} with the given data:

\{data\}

\subsection{Prompt for Simple Question Answering Generation}

You are an AI visual assistant that can analyze chart figures. You receive two detailed descriptions, raw data, and the python code drawing the figure about the same chart. The first description is the information about the raw data in the chart. The second description is about the chart figure based on Python code. In addition, raw data values within the chart is given. The python code generating the chart is given as well. Answer all questions as you are seeing the chart figure. Design a question-answer pair between you and a person asking about this chart figure. The answers should be a single word or phrase, and in a tone that a visual AI assistant is seeing the chart figure and answering the question.

Ask diverse questions and give corresponding answers. The number of questions needs to be between 3 and 20. Include questions asking about \{Characteristics\} and so on. 
Only include questions that have definite answers:(1) one can see in the chart figure that the question asks about and can answer confidently;(2) one can determine confidently from the chart figure that it is not in the chart figure.

Do not ask any question that cannot be answered confidently. The answers should be a single word or phrase.

\#Here are some examples and remember to follow their format:

\{instruction\_examples\}

\#The first description: 

\{data\_des\}

\#The second description:

\{chart\_des\}

\#The raw data: 

\{raw\_data\}

\#The code:

\{code\}

Output:

\subsection{Prompt for Complex Question Answering Generation}

You are an AI visual assistant that can analyze chart figures. You receive two detailed descriptions, raw data, and the python code drawing the figure about the same chart. The first description is the information about the raw data in the chart. The second description is about the chart figure based on Python code. In addition, raw data values within the chart is given. The python code generating the chart is given as well. Answer all questions as you are seeing the chart figure. Design a question-answer pair between you and a person asking about this chart figure. The answers should be a single word or phrase, and in a tone that a visual AI assistant is seeing the chart figure and answering the question.

Ask diverse questions and give corresponding answers. The number of questions needs to be between 2 and 10. Include questions asking about \{Characteristics\} and so on. 

Only include questions that have definite answers:(1) one can see in the chart figure that the question asks about and can answer confidently;(2) one can determine confidently from the chart figure that it is not in the chart figure.

Do not ask any question that cannot be answered confidently. You need to follow the steps below to reason before generating the answer:

(1) Describe the relevant information from the image needed to answer the question. (Do not mention the 'raw data', 'code', 'table' and so on, use 'chart' or 'figure' instead)

(2) Use the information described in (1) to reason about the problem by working step by step to arrive at the final answer.

(3) state the final answer.

\# Here are some examples and remember to follow their format: 

\{cot\_example\}

\# The first description: 

\{data\_des\}

\# The second description: 

\{chart\_des\}

\# The raw data:

\{raw\_data\}

\# The code:

\{code\}

Output:


\section{Dataset Details}

\subsection{Types of Charts}
\label{appx:types-of-charts}

9 Types of charts in SynChart:

bar chart, line chart, radar chart, stacked bar plot (stacked bar chart), doughnut chart, pie chart, scatter plot, boxplot, stacked area chart.


\subsection{Col/Row Constraints}

\begin{table}[h]
 \caption{Col/Row constraints for each type of chart.}
  \centering
  \begin{tabular}{p{3cm}p{12cm}}
    \toprule
    Chart type & Col/row constraints \\
    \midrule
     bar chart & \{random.randint(1,3)\} variables, besides the category axis, and \{random.choice([2,3,4,5,6,7,8,9,10,15,20,25,30])\} rows \\
line chartf & \{random.randint(1,2)\} columns for the x-axis (if including time) and \{random.randint(1,5)\} columns for the y-axis, with the number of rows is \{random.choice([2,3,4,5,6,7,8,9,10,15,20,25,30])\} \\
radar chart & \{random.randint(1,10)\} rows and \{random.randint(3,10)\} columns, where each column represents a variable or dimension to be plotted \\ 
stacked bar plot & \{random.randint(2,10)\} columns and \{random.choice([2,3,4,5,6,7,8,9,10,15,20,25,30])\} groups \\
doughnut chart & 1 column of data (beyond the label column) and \{random.choice([2,3,4,5,6,7,8,9,10,15,20])\} rows \\
stacked bar chart & "\{random.randint(2,10)\} columns and \{random.choice([2,3,4,5,6,7,8,9,10,15,20,25,30])\} rows \\ 
pie chart & 1 column of categorical data and 1 column of numerical data, with \{random.randint(2,8)\} rows \\
scatter plot & \{random.choice([2]*10 + [3,4,5])\} columns and \{random.choice([2,3,4,5,6,7,8,9,10,15,20,25,30])\} rows \\ 
boxplot & \{random.randint(1,10)\} columns and \{random.choice([2,3,4,5,6,7,8,9,10,15,20,25,30])\} rows \\ 
stacked area chart & \{random.randint(2,5)\} columns and \{random.choice([5,6,7,8,9,10,15,20,25,30])\} rows \\
    \bottomrule
  \end{tabular}
  \label{tab:col-row-constraints}
\end{table}

\subsection{Covered Chart Types of Chart Engines}

\begin{table}[h]
 \caption{Covered Chart Types of Chart Engines}
  \centering
  \begin{tabular}{p{3cm}p{12cm}}
    \toprule
    Chart engine & Covered chart types \\
    \midrule
    Matplotlib & All \\
    Plotly & All \\
    Seaborn & bar chart, line chart, scatter plot, boxplot \\
    Bokeh & bar chart, line chart, stacked bar plot, scatter plot, boxplot, stacked area chart \\
    \bottomrule
  \end{tabular}
  \label{tab:engine-chart-types}
\end{table}

\end{document}